\documentclass{article} 
\usepackage{iclr2027_conference,times}

\usepackage{amsmath,amsfonts,bm}

\def\eqref#1{equation~\ref{#1}}

\def\1{\bm{1}}

\DeclareMathAlphabet{\mathsfit}{\encodingdefault}{\sfdefault}{m}{sl}
\SetMathAlphabet{\mathsfit}{bold}{\encodingdefault}{\sfdefault}{bx}{n}

\usepackage{microtype}
\usepackage{hyperref}
\usepackage{url}
\usepackage{booktabs}

\usepackage{microtype}
\usepackage{graphicx}
\usepackage{subcaption}
\usepackage{booktabs} 
\usepackage{float}

\usepackage{caption}
\usepackage{multirow}
\usepackage{tabularx}
\usepackage{xurl}
\usepackage{wrapfig}
\makeatletter 
\newcommand{\needspace}[1]{\par\penalty-100\begingroup\setlength{\dimen@}{#1}\dimen@ii\pagegoal\advance\dimen@ii-\pagetotal\ifdim\dimen@>\dimen@ii\ifdim\dimen@ii>\z@\vfil\fi\break\fi\endgroup}
\makeatother
\usepackage{colortbl}

\newcommand{\name}{SkillGym}
\newcommand{\numtasks}{$6.8\mathrm{k}$}

\newcommand{\numpostrajs}{$19\mathrm{k}$}
\newcommand{\numrawskills}{$184\mathrm{k}$}
\newcommand{\numselectskills}{$11,897$}

\title{\name: Training Skill-Use Agents with Automatic Verifiable Environment Generation}

\iclrfinalcopy
\author{Renxi Wang \quad Mingshan Hee \quad Fajri Koto \quad Timothy Baldwin \quad Haonan Li \\
Mohamed bin Zayed University of Artificial Intelligence\\
\texttt{\{renxi.wang,haonan.li\}@mbzuai.ac.ae} \\
}

\begin{document}

\maketitle

\begin{abstract}
Skills equip LLM agents with professional knowledge and guidance to complete long-horizon and complex tasks. Although skills have been widely adopted in recent agent paradigms and harnesses, how to synthesize reliable training data and how to train agents for skill use remain underexplored. In this work, we propose \name, an automatic pipeline to build verifiable environments, collect trajectories, and train skill-use agents. \name\ first crawls a large volume of skills from the internet, then keeps those whose workflows can run reproducibly offline. A builder-reviewer pipeline is used to construct difficulty-controlled tasks, spanning four task types, each with a reference solution and an executable verifier. With this pipeline, we build \numtasks\ environments and collect \numpostrajs\ verified successful trajectories for supervised finetuning. Finetuning on these trajectories improves LLMs of different families and sizes, from 2B to 122B parameters across four skill-use benchmarks; Our Qwen3.5-9B SFT model outperforms the 397B untrained model on two of them. Further analysis shows that training teaches agents to invoke skills, raising the rate of reading the relevant skill from 28\% to 96\%, and that the gains hold across reasoning structures, extending to task types that form a minority of the training data and to skills held out from training. Code and data are available at \url{https://github.com/Reason-Wang/SkillGym}.
  
\end{abstract}

\section{Introduction}

Agent skills are reusable packages that provide task guidance, factual knowledge, or runnable scripts to large language model (LLM) agents to help them complete tasks \citep{zhang2025agentskills} (e.g. Anthropic's pptx and mcp skills). LLM agents integrate and discover skills at inference time. Augmenting agents with skills has been shown to improve their performance in tasks that require domain knowledge or expertise \citep{li2026skillsbench}. Because they are flexible and require no weight updates, skills are now standard in modern agent harnesses, such as Claude Code \citep{anthropic_claude_code}, Codex \citep{openai_codex_cli}, and OpenClaw \citep{openclaw}. However, effectively using skills requires the agent to interpret skill content, determine how to apply it to current task, and translate it into appropriate actions \citep{han2026skill,tan2026skt,han2026agent}.
For example, a documented workflow may require adapting its steps to the available inputs, resolving dependencies, and responding to unexpected execution outcomes. This motivates training agents to apply externally provided skills more effectively, with the aim of transferring the learned behavior to new skills and tasks.

Existing skill-use training approaches mostly derive skills from an agent's own experience in fixed environments \citep{xia2026skillrl,shi2026skill1,yang2026skillforge}, which confine skill-use learning to a few task domains. While we reverse the direction to start from skills and build environments. Public community-written skills cover many domains, such as software engineering, science, finance, document processing, and marketing \citep{skills_sh,claude_skill_registry,li2026organizing}.
However, these skills are written as reusable resources rather than training materials: none comes with a task, an environment, or a way to check success. Turning them into training data raises challenges: First, tasks must be skill-critical. Applying the skill should decide the outcome while the task stays solvable without it, so that agents learn to use the skill rather than bypass it. Second, outcomes must be verified reliably. Such verified outcomes should reflect the task requirements, recognize valid alternative solutions, and distinguish successful completions from superficially plausible outputs.

In this work, we introduce \name, an automatic agentic pipeline that transforms community-written skills into skill-critical tasks. From a curated collection of skills, \name\ builds task environments and constructs problems around the applications of the knowledge, procedures, and scripts these skills are intended for. These tasks are based on four reasoning structures: procedural execution \citep{shridhar2020alfworld}, abductive diagnosis \citep{jimenez2024swe,zhao2023abductive}, constraint satisfaction \citep{xie2024travelplanner}, and partial-order planning \citep{lin2402graph,qiao2025benchmarking}. A two-stage builder-reviewer agent system first prepares the environments and then develops task instructions, initial workspaces, reference solutions, and executable verifiers. The final task is formed by combining the builder's exploration. Execution checks establish that the reference solution completes an initially unsolved task \citep{jimenez2024swe}, while reviewer feedback helps identify and repair ambiguous requirements, information leakage, and overly restrictive or insufficient verification.

In total, we construct \numtasks\ tasks and collect \numpostrajs\ verified successful trajectories from three teacher models \citep{team2026kimi,zeng2026glm,xu2026deepseek} across four agent harnesses. Supervised finetuning (SFT) on these trajectories improves six LLMs from three families, ranging from 2B to 122B parameters on four skill-use benchmarks. The finetuned models are even competitive with much larger ones. Our 9B model outperforms Qwen3.5-397B-A17B \citep{qwen3.5} on our test set and on SkillEval \citep{tan2026skt}. To summarize, our contributions are as follows:
\begin{itemize}
    \item  We introduce \name, an automated pipeline that transforms community-written skills into executable training environments. It constructs tasks across four reasoning structures and uses a builder-reviewer agent system to refine environments, reference solutions, and outcome verifiers.
    \item We construct \numtasks\ tasks and collect \numpostrajs\ interaction trajectories across multiple agent harnesses. Supervised finetuning on the resulting data improves agents' skill-use task performance, including on tasks involving skills held out from finetuning.
    \item We find training teaches agents to consult the provided skills, raising the rate from 28\% to 96\%. Annotating tasks from three benchmarks with a shared rubric, we find that the gains hold across reasoning structures and transfer to structures that form a minority of the training data.
\end{itemize}

\section{Related Work}\label{sec:related_work}
\paragraph{Learning to Use Agent Skills} Several studies have been released to improve LLM agents' skill-use capabilities, which broadly fall into training-free and training-based methods. Training-free methods usually gather experiences from interaction trajectories, and use them to guide future exploration. Voyager~\citep{wang2023voyager} builds, retrieves, and composes an expanding library of executable skills through environment feedback. SkillWeaver~\citep{zheng2025skillweaver} explores websites and develops reusable APIs that improve web-agent interaction. AgentSkillOS~\citep{li2026organizing} organizes existing skills into a capability hierarchy and retrieves relevant skills on demand. In this work, we focus on training-based methods, which update model weights with skill-related trajectories. Skill-to-LoRA~\citep{zhang2026skill} uses synthetic trajectories to train skill-specific adapters. SAGE~\citep{wang2026reinforcement} starts from SFT and uses skill-augmented RL to improve skill creation and utilization. SkillRL~\citep{xia2026skillrl} extracts reusable knowledge into a hierarchical SkillBank and jointly evolves the skill library and agent through RL. These methods derive skills from an agent's own experience in a small set of environments or compile specific skills into model weights. \name\ instead keeps skills external and trains the general capability to use public skills, spanning 3.5k skills across 18 domains, and the learned behavior transfers to skills held out from training.


\paragraph{Task and Environment Synthesis} Automatically synthesizing tasks and environments provides a scalable source for training LLM agents. We group these studies based on what drives the synthesis pipeline. Task-driven synthesis starts from a task and then constructs the related environment. Endless Terminals~\citep{gandhi2026endless} generates terminal-related task descriptions, then builds their environment container and refines it iteratively. CLI-Universe~\citep{hua2026cli} starts with three task dimensions, creates task candidates and refines them iteratively. Environment-driven synthesis starts from the environments, tools, or states~\citep{song2026envscaler, wang2026agent, dong2026agent}. EnvScaler~\citep{song2026envscaler} collects diverse environment themes, then uses LLMs to enrich environment descriptions and construct environment states. Agent-World~\citep{dong2026agent} collects thousands of real-world environment themes, then uses a deep-search pipeline to mine databases and executable tool interfaces. Tasks are synthesized on top of these environments. Skill-driven synthesis creates tasks from skills. SKT~\citep{tan2026skt} synthesizes template-driven task packages from skills.sh skills with difficulty control and verified trajectories, and trains agents on 4k such tasks across two harnesses. \name\ is also skill-driven, but builds tasks around explicit reasoning structures and pairs execution checks with a reviewer that audits verifiers for overly strict or insufficient checks and information leakage. The resulting tasks cover all five reasoning structures we annotate, whereas SKT's evaluation set concentrates on applying skill-provided rules (Section~\ref{sec:task_structures}).

\paragraph{Benchmarking Agent Skill Use} Existing benchmarks mainly evaluate LLM agent skill-use capabilities through task completion. SkillsBench~\citep{li2026skillsbench} collects expert-written tasks that require skills, span multiple domains, and are all verifiable by deterministic checks. SWE-Skill-Bench~\citep{han2026agent} curates skills for SWE-style tasks, studying LLM agents on real software repositories with execution-based tests. AgentSkillOS~\citep{li2026organizing} evaluates skill retrieval and orchestration through pairwise assessment of generated artifacts. Skill-Use-Bench~\citep{han2026skill} decomposes skill use into triggering, procedural compliance, and boundary adherence, scoring trajectories under progressive disclosure. These benchmarks contain at most a few hundred tasks and are designed for evaluation. \name\ instead provides thousands of verifiable tasks for training, and we use these benchmarks to measure transfer (Section~\ref{sec:experiments}); our structure annotation further shows that they place different demands on agents (Section~\ref{sec:task_structures}).

\section{\name}\label{sec:skill_gym}
\begin{figure}
\centering
\includegraphics[width=\textwidth]{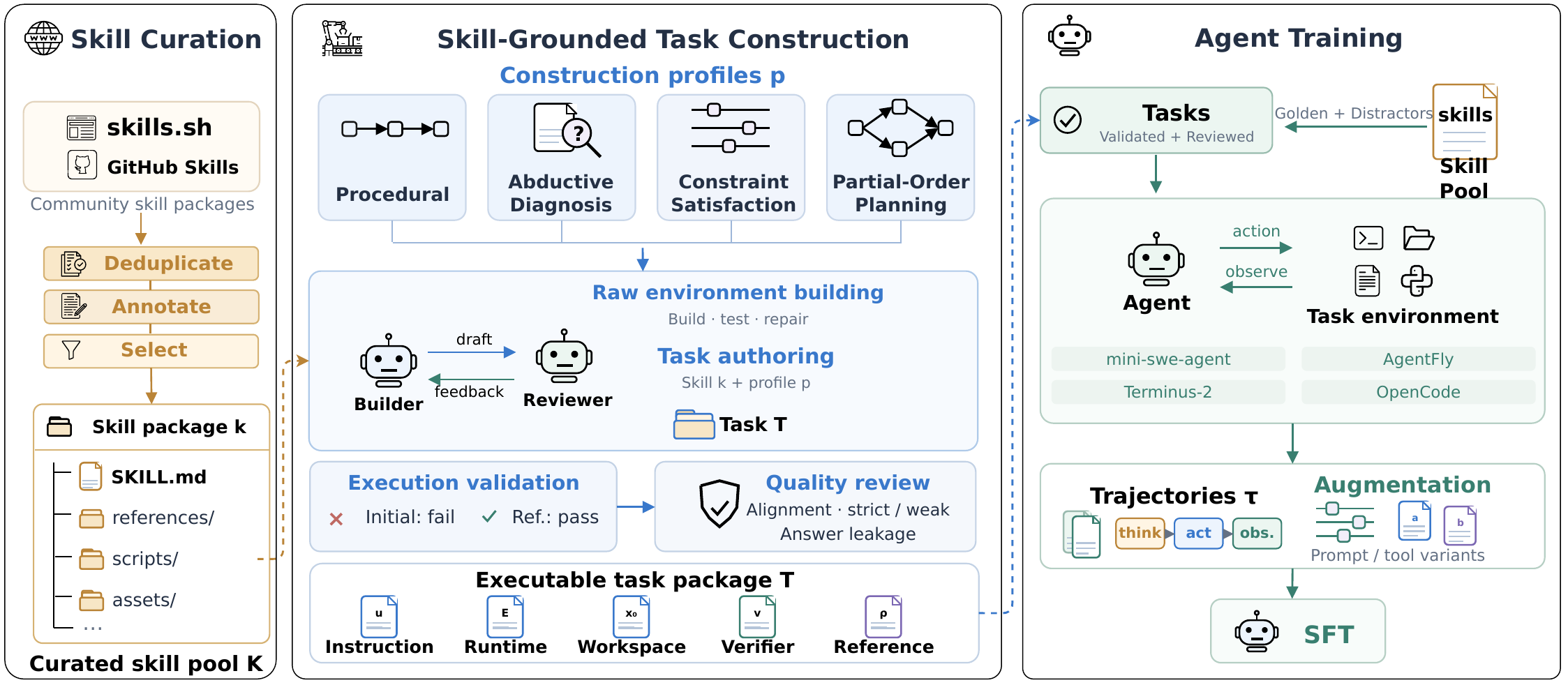}
\caption{Overview of \name. Public skill packages are curated and transformed into executable tasks through two stages of builder-reviewer interaction: raw environment building and task authoring. Execution validation and quality review support task refinement, followed by multi-harness trajectory collection and supervised finetuning.}
\label{fig:skillgym_method}
\end{figure}

\name\ converts public skills into executable task environments with a three-stage pipeline: (I) collecting and curating skill packages; (II) constructing skill-critical tasks; and (III) collecting trajectories for agent training. Figure~\ref{fig:skillgym_method} gives an overview of the pipeline.


\subsection{Skill Collection and Curation}


A skill can serve as the basis of a task only if it is (i) complete and substantive, with non-empty documentation and all referenced files available, (ii) executable in an isolated container without internet access, GPU requirements, or interactive input, and (iii) sufficiently clear to support the construction of meaningful workflow tasks.



\paragraph{Skill Crawling} Each skill is downloaded and stored as a single folder, where a \texttt{SKILL.md} file must exist to specify basic skill information, such as its name, description, and body. We collect skills from two main sources. The first is \texttt{skills.sh}\footnote{\url{https://www.skills.sh}}, from which we collect around $9.7$k top-ranked skills that represent the most widely used skills in the community. The second is claude-skill-registry\footnote{\url{https://github.com/majiayu000/claude-skill-registry}}, a public aggregation of agent skills from GitHub. We download all files from skills' original sources to ensure every skill is complete. We crawl \numrawskills\ skill entries from the two sources, of which 51k unique skills can be fetched after deduplication; Table~\ref{tab:stats_funnel} in Appendix~\ref{app:statistics} lists the counts per source and stage.


\paragraph{Skill Annotation} 

Each skill is annotated on three dimensions using a combination of rule-based checks and LLM-based annotation: (i) basic properties, covering package completeness, language, file count, and folder size; (ii) runtime requirements, covering network access, GPU requirements, and dependencies; and (iii) quality, covering the coherence of the skill description and clarity of its requirements. A detailed description of the annotation properties is provided in Appendix~\ref{app:annotation}. To ensure annotation quality, a human expert independently annotates 100 sampled skills, resulting in an agreement of $94\%$ with the automatic annotations.


\paragraph{Skill Filtering \& Selection} 

Skills are first filtered by basic requirements, retaining only valid, non-empty, and coherently written skills in English. The remaining skills are selected for task creation based on whether they can support stable execution with modest resources. Specifically, selected skills must operate without runtime network access or GPUs, avoid destructive actions such as writing outside the working directory, and remain within a file-count limit of at most 300 files. The complete selection criteria are provided in Table~\ref{tab:skill_annotation}. This process yields a curated pool $\mathcal{K}$ of \numselectskills\ skills, whose domain distribution is shown in Figure~\ref{fig:profile_example}b and Table~\ref{tab:stats_domains}.


\subsection{Task Creation}

\subsubsection{Design Principle}\label{sec:design}
Task construction follows two main principles. First, each task must be skill-critical. Completing it requires applying a procedure or utility provided by the skill, with at least one consequential step relying on skill-specific knowledge not stated in the task instruction. The task should remain solvable through inspection and experimentation, but access to the skill should provide a clear advantage. Second, task outcomes must be reliably verifiable. Each task therefore includes a reference solution and an executable verifier that checks the required outcome while allowing alternative valid solutions. Each finalized task is represented as $T=(u,\mathcal{E},x_0,v,\rho)$, consisting of a task instruction $u$, an executable environment $\mathcal{E}$, an initial workspace state $x_0$, an executable verifier $v$, and a reference solution $\rho$.


\subsubsection{Task Profiles}

Prior work has explored a range of reasoning structures, including multi-hop tasks solved step by step \citep{shi2026taskcraft,fan2026toward,tao2026webshaper}, diagnosis of faulty systems \citep{jimenez2024swe,zhao2023abductive}, planning under competing constraints \citep{xie2024travelplanner}, and plans whose steps are only partially ordered \citep{lin2402graph,qiao2025benchmarking}. Each of these studies centers on a single structure, and to our knowledge none combines several structures to synthesize verifiable, skill-grounded tasks. We therefore define four task profiles, one for each reasoning structure, to diversify how skills are applied.

\begin{figure}[t]
\centering
\includegraphics[width=\textwidth]{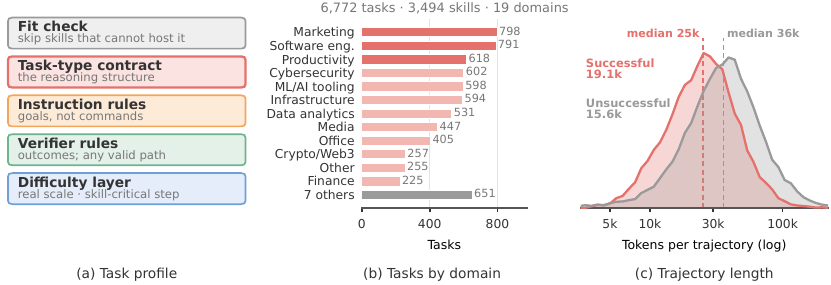}
\caption{\textbf{(a)} The parts shared by every task profile; the task-type contract differs across the four reasoning structures. \textbf{(b)} Tasks per domain (6,772 tasks from 3,494 skills). \textbf{(c)} Token length of successful (19.1k) and unsuccessful (15.6k) teacher trajectories; dashed lines mark the medians.}
\label{fig:profile_example}
\end{figure}

\begin{itemize}
    \item \textbf{Procedural.} The task consists of a sequence of dependent steps, where each step uses the artifact produced by the previous step. The verifier checks the result of each step.
    \item \textbf{Abductive.} The task starts from a system with incorrect observed behavior. The agent must identify the unstated cause, repair the system, and demonstrate the corrected behavior. The instruction provides the symptoms but does not reveal the cause.
    \item \textbf{Constraint satisfaction.} The task requires a deliverable that satisfies multiple measurable constraints. Since satisfying one constraint may violate another, the agent must measure the results and iteratively refine the solution.
    \item \textbf{Partial order.} The task defines dependencies as a directed acyclic graph sampled for each task. This includes joins where outputs from different branches must agree and inputs that are reused later and therefore cannot be modified in place. The agent must determine a valid execution order and produce all required deliverables.
\end{itemize}
Each task profile is a prompt block given to the builder agent, and Figure~\ref{fig:profile_example}a shows its parts. Besides the task-type contract above, it contains a \emph{difficulty layer} that makes skill-specific knowledge important for solving the task. It requires realistic input scales that prevent solutions from being easily computed or guessed, value-based verification that recomputes expected values from the inputs rather than hard-coding them, and at least one consequential step that depends on non-obvious skill-specific knowledge left unstated in the instruction. Depending on the profile, this step involves a documented edge case, the hidden cause of a failure, an adversarial constraint, or a join that produces an incorrect result by default. Each profile also carries a fit check and rules for the instruction and verifier; Section~\ref{sec:construction} describes how the builder-reviewer loop enforces them. Appendix~\ref{app:profiles} gives the full prompts of the four profiles, and Appendix~\ref{app:examples} shows an example task for each.

\subsubsection{Two-Stage Agentic Construction}\label{sec:construction}
\paragraph{Environment Building} 

Basic environments are constructed with all required dependencies installed using a builder-reviewer agentic system. The builder creates a Dockerfile and any required dependency files based on the skill requirements (if specified). The reviewer builds the image, launches the container, and independently validates the environment by running relevant tests and commands. If all checks pass, the Dockerfile and dependency files are accepted and retained as the final deliverables. Otherwise, the reviewer provides feedback to the builder, which revises the environment accordingly. This process repeats until the environment passes validation or the maximum number of iterations is reached.

\paragraph{Task Construction} 

Task construction reuses the builder-reviewer system with three gates. First, the builder applies the profile's \emph{fit-check gate} and skips the skill if it cannot support the target task type; otherwise, it builds a task package containing a task instruction, initial workspace files, a reference solution script, and a verifier script. Second, a \emph{validity gate} executes the package and accepts it only if the verifier fails on the initial workspace $x_0$ and passes on the workspace produced by the reference solution $\rho$ in environment $\mathcal E$:

$$
  v(x_0)=0,
  \qquad
  v\left(\operatorname{Exec}(\rho;\mathcal E,x_0)\right)=1.
$$

Third, at the \emph{quality gate}, the reviewer checks the package against the profile's instruction and verifier rules: it looks for information leakage among the skill, instruction, and verifier, and for checks that are too strict to accept valid alternative solutions or too weak to reject incorrect ones. Tasks that pass all gates are accepted; otherwise, the reviewer returns feedback to the builder, and the loop repeats until the task passes or the maximum number of iterations is reached. Appendix~\ref{app:builder_prompts} gives the prompts of the builder and reviewer agents in both stages.

\subsection{Trajectory Collection}

Training trajectories are generated by three open-weight LLMs, Kimi-K3, DeepSeek-V4-Flash, and GLM-5.2, using four agent harnesses, MiniSwe-Agent, AgentFly, Terminus-2, and OpenCode. Using multiple LLMs captures variations in reasoning, action sequences, and tool-use behaviors, while reducing dependence on a single model. Using multiple harnesses further exposes the models to different interaction interfaces and tool-use formats: MiniSwe-Agent uses Bash commands for all agent actions, AgentFly and OpenCode provide dedicated tools for file operations and command execution, while Terminus-2 uses a JSON-based protocol for command execution. Further diversity is introduced by varying the system prompts, tool names, and tool schemas. Executing an agent within a task environment produces a trajectory $\tau=(o_0,a_0,\ldots,o_L)$, consisting of observations $o_t$ and actions $a_t$.


\section{Experiments}\label{sec:experiments}

\begin{table}[t]
\centering
\caption{Main evaluation results, higher is better. SkillEval and SkillsBench show mean $\pm$ standard deviation over three runs; the other benchmarks use one run. Bold marks the higher score within each pair where both results are available. $^{\diamond}$Teacher models whose trajectories form our training data. More reference models are in Appendix~\ref{app:reference_models}. $^{\S}$Reported by \citet{tan2026skt} from their paper.}
\label{tab:main_experiments}
\begingroup
\small
\setlength{\tabcolsep}{3pt}
\renewcommand{\arraystretch}{1.15}
\begin{tabular*}{\textwidth}{@{\extracolsep{\fill}}lrcccc@{}}
\toprule
\textbf{Model} & \textbf{Size} & \textbf{\name} & \textbf{SkillEval} & \textbf{SkillsBench} & \textbf{Skill-Use-Bench} \\
\midrule
Qwen3.5 & 397B/17B & $55.8$ & $73.8 \pm 0.8$ & $33.2 \pm 3.0$ & $23.0$ \\
GLM-5.2$^{\diamond}$ & 753B & $66.5$ & $79.6 \pm 0.8$ & $57.8 \pm 4.8$ & $76.3$ \\
DeepSeek-V4-Flash$^{\diamond}$ & 284B/13B & $61.5$ & $78.6 \pm 0.2$ & $58.7 \pm 1.0$ & $56.1$ \\
Kimi-K3$^{\diamond}$ & 2.8T/104B & $68.8$ & $81.0 \pm 0.9$ & $53.5 \pm 4.8$ & $76.3$ \\
\midrule
\multicolumn{6}{@{}l}{\textit{Base and \name\ SFT comparisons}} \\[2pt]
MiniCPM5 & 2B & $33.2$ & $63.0 \pm 0.6$ & $\mathbf{10.8 \pm 2.7}$ & $24.1$ \\
\quad + \name\ SFT & 2B & $33.2$ & $\mathbf{68.1 \pm 2.0}$ & $6.1 \pm 0.8$ & $\mathbf{44.7}$ \\
\addlinespace[3pt]
Ministral-3 & 8B & $17.5$ & $55.7 \pm 3.1$ & $3.9 \pm 2.7$ & $4.3$ \\
\quad + \name\ SFT & 8B & $\mathbf{49.5}$ & $\mathbf{77.5 \pm 0.4}$ & $\mathbf{16.9 \pm 1.3}$ & $\mathbf{55.1}$ \\
\addlinespace[3pt]
Qwen3.5 & 4B & $33.8$ & $62.8 \pm 1.5$ & $10.1 \pm 1.3$ & $8.8$ \\
\quad + \name\ SFT & 4B & $\mathbf{47.0}$ & $\mathbf{70.8 \pm 1.6}$ & $\mathbf{14.3 \pm 1.0}^{\dagger}$ & $\mathbf{48.7}$ \\
\addlinespace[3pt]
Qwen3.5 & 9B & $41.3$ & $65.7 \pm 1.1$ & $14.8 \pm 3.5$ & $14.8$ \\
\quad + SKT SFT$^{\S}$ & 9B & \textemdash & $74.1 \pm 1.4$ & $14.2 \pm 0.6$ & \textemdash \\
\quad + \name\ SFT & 9B & $\mathbf{59.5}$ & $\mathbf{74.7 \pm 1.2}$ & $\mathbf{22.4 \pm 4.2}$ & $\mathbf{49.6}$ \\
\addlinespace[3pt]
Qwen3.5 & 27B & $55.3$ & $76.2 \pm 1.1$ & $32.6 \pm 2.3$ & $28.3$ \\
\quad + \name\ SFT & 27B & $\mathbf{62.8}$ & $\mathbf{80.1 \pm 1.5}$ & $\mathbf{47.4 \pm 5.1}^{\dagger}$ & $\mathbf{74.2}$ \\
\addlinespace[3pt]
Qwen3.5 & 122B/10B & $53.0$ & $69.8 \pm 1.1$ & $30.1 \pm 2.5$ & $16.8$ \\
\quad + \name\ SFT & 122B/10B & $\mathbf{65.0}$ & $\mathbf{80.0 \pm 0.3}$ & $\mathbf{53.6 \pm 5.2}$ & $\mathbf{72.2}$ \\
\bottomrule
\end{tabular*}
\par
\endgroup
\end{table}

\subsection{Setup} \label{sec:experiments_setup}
\paragraph{Training} For SFT, we use our collected trajectories to train LLMs for 2 epochs. Learning rate is set to $10^{-5}$, with a linear scheduler decaying to zero and AdamW optimize to update weights. We use 128 as the batch size, and train the model for 64 GPU hours. For LLMs, we select MiniCPM5-2B \citep{minicpm4}, Ministral-3-8B \citep{liu2026ministral}, and Qwen3.5 series \citep{qwen3.5}, including 4B, 9B, 27B, and 122B-A10B.

\paragraph{Evaluation} For main experiments, we evaluate models on (I) \name\ splited test set, which contain a held-in and held-out subset, held-in consists of unseen tasks with seen skills during training, while held-out consists of unseen tasks with unseen skills. (II) SkillEval \citep{tan2026skt} is a synthesitic dataset constructed by SKT, aother automatic agent pipeline, consisting of single and multiple-skill tasks; (III) SkillsBench \citep{li2026skillsbench}, a general skill-use benchmarks with all samples crurated by human experts spanning 8 domains; (IV) Skills-Use-Bench \citep{han2026skill}, which measures the agent in three dimensions: whether invokes the relevant skill, whether faithfully follows prescribed procedure and whether it avoids forbidden operations. We report overall task success on \name, mean normalized reward on SkillEval and SkillsBench, and the SU score on Skill-Use-Bench, which combines the three skill-use dimensions.

We use MiniSwe-Agent \citep{yang2024sweagent} as the evaluation harness, which allows only bash tool for agent to use. Skills' names and descriptions are put in the system prompt, while detailed contents need to be disclosed by the agent itself.

\subsection{Results}
Table~\ref{tab:main_experiments} compares six backbones from three families, from 2B to 122B parameters, before and after \name\ SFT. Training improves the model in 22 of the 24 comparisons, with average gains of 13.8 points on the \name\ test set, 9.7 on SkillEval, 9.7 on SkillsBench, and 41.2 in Skill-Use-Bench SU. The largest and most uniform gain is in skill use itself: SU rises by 21 to 55 points for every backbone, and Section~\ref{sec:behavior_analysis} analyzes this change. Gains on the \name\ test set and SkillEval are largest for Ministral-3, the weakest base model, whereas gains on SkillsBench, whose human-written tasks are the hardest, grow with model size within the Qwen3.5 family, from +4.2 at 4B to +23.5 at 122B; the 2B model is the only one whose SkillsBench score drops. The trained models also compare well with much larger ones: the 9B model outperforms Qwen3.5-397B-A17B on the \name\ test set and SkillEval, the 27B model exceeds two of its three teachers on SkillEval, and our 9B model matches the SkillEval score reported for SKT without using its data, although the two use different harnesses.




\begin{figure}[t]
\centering
\includegraphics[width=\textwidth]{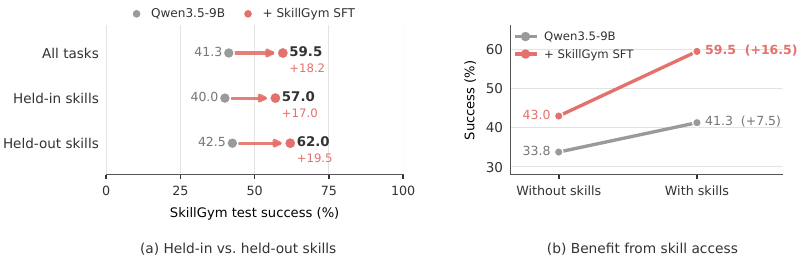}
\caption{(a) Qwen3.5-9B success rate before and after \name\ SFT, overall (400 tasks) and per skill split (200 each), with skills available. (b) Overall success rate without and with skills.}
\label{fig:skillgym_held_in_out}
\label{fig:skillgym_skill_access}
\end{figure}

\noindent\textbf{Generalization to held-out skills.}
Figure~\ref{fig:skillgym_held_in_out}a separates the \name\ results by skill split. SFT increases success from 40.0\% to 57.0\% on held-in skills and from 42.5\% to 62.0\% on held-out skills, 17.0 and 19.5 percentage points respectively. The gains on both splits show that the improvement extends to skills excluded from training. On these evaluation sets, held-out success is higher than held-in success for both models, with a difference of 2.5 percentage points for the base model and 5.0 points for SFT.

\noindent\textbf{Benefit from skill access.}
Figure~\ref{fig:skillgym_held_in_out}b compares overall success on the \name\ test set with and without skills. Providing skills increases base-model success from 33.8\% to 41.3\%, a gain of 7.5 percentage points. For the SFT model, success increases from 43.0\% to 59.5\%, a gain of 16.5 percentage points. SFT thus improves success even without skills, by 9.2 points, and more than doubles the benefit that the model draws from skill access. The same holds on SkillsBench, where the SFT model scores 9.4 without skills and 22.4 with them. The training therefore does not only strengthen general task-solving ability, it teaches the agent to make better use of the skills it is given.

\section{Analysis}\label{sec:analysis}
\subsection{Training improves model's beharior}
\label{sec:behavior_analysis}

\noindent\begin{minipage}{\textwidth}
\begin{wrapfigure}{r}{0.46\textwidth}
\centering
\includegraphics[width=\linewidth]{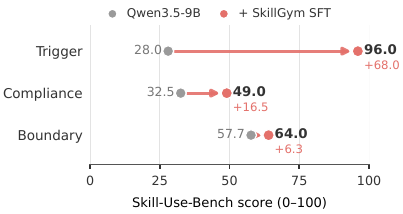}
\caption{Mean component scores on Skill-Use-Bench for Qwen3.5-9B.}
\label{fig:skill_use_components}
\end{wrapfigure}
Figure~\ref{fig:skill_use_components} breaks the Skill-Use-Bench score of Qwen3.5-9B into its three components, refer to Section~\ref{sec:experiments_setup} for definitions. All three improve after SFT. The largest change is in Trigger, which rises from 28.0 to 96.0, showing that the base model opens the relevant skill in fewer than a third of tasks, the main reason its skill-use score is low, whereas the trained model consults it almost always. Consulting the skill is a prerequisite for applying it, and the trained model also follows what it reads more faithfully. The Compliance score rises from 32.5 to 49.0 and Boundary from 57.7 to 64.0. The consulted skills are also put to use. With skills available, the trained model gains more than twice as much success on the \name\ test set as the base model as shown in Figure~\ref{fig:skillgym_held_in_out}b.
\par
\ifnum\value{WF@wrappedlines}>1
  \vspace{\dimexpr\value{WF@wrappedlines}\baselineskip-\baselineskip\relax}
\fi
\WFclear
\end{minipage}\par

\subsection{Task Structures}
\label{sec:task_structures}

\noindent\begin{minipage}{\textwidth}
\begin{wrapfigure}{r}{0.44\textwidth}
\centering
\vspace{-6pt}
\includegraphics[width=\linewidth]{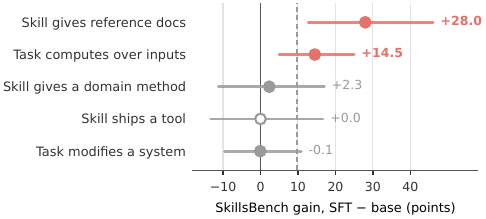}
\caption{SkillsBench gains by what the skill provides and what the task asks. Lines show 95\% task-bootstrap intervals; the dashed line is the overall gain; the hollow marker has fewer than 15 tasks.}
\label{fig:skillsbench_by_role}
\end{wrapfigure}
To test whether the gains merely reflect agreement between training and test distributions, we annotate tasks from all three benchmarks with a shared rubric of five reasoning structures, annotation details are in Appendix~\ref{app:task_structure}. The benchmarks differ markedly, as shown in Figure~\ref{fig:task_composition}. Nearly all SkillEval tasks apply skill-provided rules to a list of items, while SkillsBench relies more on skills that supply domain methods or reference documentation. Yet SFT improves every structure by similar amounts, by 16.9 to 22.1 points on \name\ and 15.4 to 16.4 on SkillEval, and a regression controlling for task difficulty finds no structure-specific effect as shown in Figure~\ref{fig:gains_by_structure}. Gains also reach structures that are rare in training: rule application is the primary structure of only 12\% of training tasks, yet SkillEval's rule-application tasks improve by 13.9 points. Transfer is weakest where a skill must be \emph{applied}. As Figure~\ref{fig:skillsbench_by_role} shows, on SkillsBench, tasks whose skills provide reference documentation gain 28.0 points, but those requiring a domain method or a shipped tool, or modifying an existing system, do not improve. Together with Section~\ref{sec:behavior_analysis}, this suggests that training teaches agents to consult skills more than to apply their methods, pointing to method and tool-centric tasks as a target for data construction.
\par
\ifnum\value{WF@wrappedlines}>1
  \vspace{\dimexpr\value{WF@wrappedlines}\baselineskip-\baselineskip\relax}
\fi
\WFclear
\end{minipage}\par

\subsection{Data Ablations}
\label{sec:ablations}

\begin{table}[t]
\centering
\small
\setlength{\tabcolsep}{4.5pt}
\caption{Data ablations on Qwen3.5-9B (64k context, 2 epochs). Each pair is trained on a token-matched subset of \name{} trajectories (110M tokens for review, 132M for structure).}
\label{tab:ablations}
\begin{tabular}{lcccccc}
\toprule
 & Ours & SkillsBench & \multicolumn{2}{c}{SkillEval} & \multicolumn{2}{c}{Skill-Use-Bench} \\
\cmidrule(lr){4-5}\cmidrule(lr){6-7}
Training data & & & Pass & Strict & SU & Compl. \\
\midrule
Qwen3.5-9B (no SFT) & 41.3 & 14.8 & 65.7 & 17.3 & 14.8 & 51.4 \\
\name{} (full) & 59.5 & 22.4 & 74.7 & 33.3 & 49.6 & 46.9 \\
\midrule
Validated only & 40.5 & 10.8 & 41.2 & 18.3 & 45.1 & 43.1 \\
Review-approved & \textbf{45.2} & \textbf{12.0} & \textbf{47.9} & \textbf{24.7} & \textbf{47.4} & \textbf{52.4} \\
\midrule
Procedural only & \textbf{42.5} & \textbf{17.4} & \textbf{56.0} & 24.7 & \textbf{52.2} & \textbf{56.2} \\
Four-profile mix & 39.4 & 13.2 & 55.0 & \textbf{26.3} & 50.1 & 55.4 \\
\bottomrule
\end{tabular}
\end{table}

We ablate two design choices on token-matched subsets of the training data (Table~\ref{tab:ablations}). \textbf{Quality review.} We compare training on trajectories from \emph{review-approved} tasks, which passed both the validity and quality gates (Section~\ref{sec:construction}), with trajectories from tasks that were \emph{validated only}: they pass the validity gate, but the reviewer did not approve them within its round budget. At the same token budget, training on review-approved tasks outperforms training on validated-only tasks on every metric, by 4.7 points on our test set, 6.7 on SkillEval, and 9.3 on Skill-Use-Bench completion, indicating that quality review improves the training signal beyond execution validation. \textbf{Task structure.} Training on a mix of all four task profiles performs on par with training on procedural tasks alone at the same token budget: the two are within run-to-run variation on SkillEval and Skill-Use-Bench, and the mix is slightly lower on SkillsBench. Together with the uniform gains across reasoning structures (Section~\ref{sec:task_structures}), this suggests that the skill-use behavior acquired in training does not depend on matching the structure of training and test tasks; at this scale, the additional profiles broaden coverage rather than raise average scores. Both subsets are much smaller than the full data, and all ablation models fall below the full \name{} model; on SkillEval they fall below the base model because many rollouts end in repeated reasoning that exhausts the output budget. The ablations should therefore be read as comparisons within each pair.

\needspace{8\baselineskip}
\subsection{Are the Tasks Skill-Critical?}
\label{sec:skill_critical}

\begin{wraptable}{r}{0.42\textwidth}
\centering
\small
\vspace{-12pt}
\caption{Kimi-K3 on the \name\ test set without and with skills.}
\label{tab:skill_critical}
\setlength{\tabcolsep}{4pt}
\begin{tabular}{lccc}
\toprule
Task type & w/o & w/ & Gain \\
\midrule
Procedural & 31.0 & 47.0 & +16.0 \\
Constraint sat. & 79.8 & 89.9 & +10.1 \\
Abductive & 65.7 & 69.7 & +4.0 \\
Partial order & 65.7 & 68.7 & +3.0 \\
\midrule
All & 60.5 & 68.8 & +8.3 \\
\bottomrule
\end{tabular}
\vspace{-6pt}
\end{wraptable}
The design principle in Section~\ref{sec:design} requires tasks that become easier with the skill but remain solvable without it. We test this with Kimi-K3, one of the strongest available models and one of our teachers, which we evaluate on the \name\ test set with and without skills, as reported in Table~\ref{tab:skill_critical}. Even this model benefits from skills: success rises from 60.5\% to 68.8\%. The effect is consistent across tasks: 52 tasks are solved only with the skill and 19 only without it (McNemar exact test, $p\approx10^{-4}$). At the same time, 60.5\% of tasks are solved without skills, so the skills make tasks easier rather than gating them. The benefit is largest for procedural tasks (+16.0), whose solution follows the skill's workflow, and smallest for abductive and partial-order tasks (+4.0 and +3.0), where a strong model can often recover the needed knowledge by inspection and experimentation. Because Kimi-K3 also served as a teacher, these tasks are not adversarial to it; that it still gains from skills indicates that the tasks encode skill-specific knowledge that even a strong model does not reliably possess.

\section{Conclusion}
\label{sec:conclusion}

We presented \name, an automatic pipeline that turns community-written skills into executable, verifiable training tasks. Starting from 184k crawled skills, a builder-reviewer agent system constructs \numtasks\ skill-critical tasks across four reasoning structures, each with a reference solution and an outcome-based verifier, from which we collect \numpostrajs\ successful trajectories with four agent harnesses. Supervised finetuning on these trajectories improves LLMs of different families and sizes on four skill-use benchmarks, and the gains extend to skills held out from training. The largest behavioral change is that trained agents consult the skills they are given, and training on review-approved tasks provides a stronger signal than training on validated-only ones.
Two directions remain open. Gains on SkillsBench concentrate on skills that supply reference documentation, while skills built around a domain method or a bundled tool improve little. We also train only with supervised finetuning; reinforcement learning on \name\ environments, whose verifiers already provide outcome rewards, is a natural next step.



\newpage
\bibliography{iclr2027_conference}
\bibliographystyle{iclr2027_conference}

\appendix

\section{Skill Annotation}\label{app:annotation}

Table~\ref{tab:skill_annotation} lists the annotated properties. An LLM annotator reads each skill's \texttt{SKILL.md} and returns labels from closed vocabularies; replies outside the vocabulary are rejected and re-requested. Fields are grouped into four levels, from surface properties to semantic judgments. The LLM screening labels cover all 51,131 deduplicated skills; skills from archived repositories or bulk publishers are then removed by rule, and the identity, runtime, and quality levels are annotated on the remaining 27,569 skills. Applying the selection rules in the last column yields the \numselectskills\ skills used for task creation. Screening, runtime, and quality labels use GPT-5.4; identity labels use GPT-5.5.

\begin{table}[h]
\centering
\caption{Skill annotation fields, their values (share of annotated skills, \%), and the rule used to select skills for task creation. Fields marked $^{\dagger}$ allow several values per skill, so shares can exceed 100\%. Fields marked $^{\ddagger}$ are rule-based, computed from repository metadata and the file listing; all others are LLM labels. ``--'' means the field is recorded but not used for selection.}
\label{tab:skill_annotation}
\scriptsize
\setlength{\tabcolsep}{4pt}
\renewcommand{\arraystretch}{1.12}
\begin{tabularx}{\textwidth}{@{}l l X l@{}}
\toprule
\textbf{Level} & \textbf{Field} & \textbf{Values (\%)} & \textbf{Selection} \\
\midrule
\multirow{4}{*}{\shortstack[l]{(i) Screening\\($n{=}51{,}131$)}}
 & Label & real skill 88.9, unclear 5.5, meta/doc 3.3, template 1.0, test/demo 0.7, placeholder 0.5 & real skill \\
 & Language & English 92.4, Chinese 3.0, Japanese 1.8, Korean 0.9, mixed 0.9, other 1.0 & English \\
 & Repository$^{\ddagger}$ & archived flag, bulk-publisher flag, license, stars & not archived or bulk \\
 & Size$^{\ddagger}$ & number of files in the skill folder & $\leq$300 \\
\midrule
\multirow{3}{*}{\shortstack[l]{(ii) Identity\\($n{=}27{,}569$)}}
 & Domain & 21 domains; software engineering 53.0, agents/meta 8.8, productivity 3.9, infrastructure 3.9, writing 3.8, media 3.7, \ldots & -- \\
 & Task type$^{\dagger}$ & generate 51.7, validate 48.4, analyze 46.8, integrate 30.7, plan 26.7, orchestrate 25.2, transform 18.4, \ldots & -- \\
 & Resources$^{\ddagger}$ & presence of scripts, references, and assets & -- \\
\midrule
\multirow{9}{*}{\shortstack[l]{(iii) Runtime\\($n{=}27{,}569$)}}
 & Network access & none 63.8, runtime 31.4, setup only 4.9 & none, setup only \\
 & GPU required & no 99.9, yes 0.1 & no \\
 & Interactive & no 74.3, yes 25.7 & no \\
 & Destructive & no 85.8, yes 14.2 & no \\
 & Credentials & none 75.9, required 15.5, optional 8.6 & -- \\
 & Verification & programmatic 39.2, none 32.2, LLM/human judge 28.6 & -- \\
 & Output kind$^{\dagger}$ & stdout 48.4, files 40.3, none 31.6, external state 13.1 & -- \\
 & Runtimes$^{\dagger}$ & none 55.6, bash 31.7, python 15.6, node 8.9, other $<$1 each & -- \\
 & Services, packages & free-form lists of external services and system packages & -- \\
\midrule
\multirow{5}{*}{\shortstack[l]{(iv) Quality\\($n{=}27{,}554$)}}
 & Procedurality & procedural 58.4, mixed 38.0, declarative 3.3, persona 0.2 & not persona \\
 & Specificity & moderate 60.2, specialized 33.6, generic 6.2 & -- \\
 & Coherence & coherent 66.7, minor issues 31.9, broken 1.3 & not broken \\
 & Task scope & class of tasks 53.5, narrow 45.7, single instance 0.8 & not single instance \\
 & Abstraction & balanced 80.8, adaptable 16.7, brittle templates 2.5 & -- \\
\bottomrule
\end{tabularx}
\end{table}

\section{Task Profile Prompts}\label{app:profiles}

Each task profile is a prompt block inserted into the builder agent's system prompt; a matching addendum is appended to the reviewer's prompt so that the reviewer judges the task by the same contract. The boxes below condense the prompts for the four profiles used in \name; quoted phrases follow the prompt wording, and implementation details (tool names, file layout, optional difficulty dials) are omitted. Box~\ref{box:shared} lists the rules shared by all profiles; Boxes~\ref{box:procedural}--\ref{box:partial} give each profile's fit check, task-type contract, and difficulty layer.

\newcommand{\promptbox}[3]{%
\par\medskip\noindent\refstepcounter{promptboxctr}\label{#2}%
\fbox{\begin{minipage}{0.97\textwidth}\scriptsize
\textbf{Box~\thepromptboxctr: #1}\\[3pt]
#3
\end{minipage}}\par\medskip}
\newcounter{promptboxctr}

\promptbox{Rules shared by all profiles}{box:shared}{%
\textbf{Instruction.} \emph{``instruction.md must read like a real task a user would give''}: natural prose stating the goal and the required deliverables, named by relative path. Banned: a hints/notes/tips section; exact commands, subcommands, flags, or flag values; parameter values the task does not require; meta-commentary about the task's structure; pointers to the skill; absolute or harness paths.\\[2pt]
\textbf{Skill requirement.} \emph{``If a competent agent could complete the task from instruction.md alone without using the skill's specific knowledge \ldots\ the task is too easy.''} The exact commands live only in the reference solution.\\[2pt]
\textbf{Verifier.} Check the outcome the instruction requires, never the incidental form of the reference solution: prefer running or importing the produced artifact and asserting on what it does or outputs, and never grep the solver's source for identifiers. \emph{``test.sh must accept any correct solution.''} Do not assert names, import styles, formatting, or values the instruction leaves to the solver, and never let a check contradict a choice the instruction grants.\\[2pt]
\textbf{Validity gate.} The untouched workspace must fail the verifier and the reference solution must pass it; all checked outputs must be deterministic (sorted collections, fixed ordering, seeded randomness, pinned number formats).\\[2pt]
\textbf{Difficulty layer (common core).} (1) \emph{Realistic scale}: inputs large enough that the answer cannot be computed by hand or guessed. (2) \emph{Recompute-verify}: the verifier derives every expected value from the inputs with an independent reference implementation; \emph{``If the inputs were regenerated with different values, would test.sh still compute the correct expected answer with no edits?''} (3) \emph{Skill-sourced edge case}: plant inputs that trigger a caveat the skill documents. (4) \emph{Skill-critical, underspecified step}: the correct result hinges on a non-obvious, skill-specific fact that the instruction leaves unstated; \emph{``with the skill the fix is direct; without it the solver must research or experiment to recover the fact---harder, but still possible.''}}

\promptbox{Procedural}{box:procedural}{%
\textbf{Fit check.} If the skill cannot honestly support the required number of distinct dependent steps, skip the skill rather than pad.\\[2pt]
\textbf{Task-type contract.} \emph{``Design the task as a linear sequence of ordered steps that exercise the skill's workflow.''} The steps form a true dependency chain: \emph{``Step $k$ must consume the artifact produced by step $k{-}1$''} (raw $\rightarrow$ A $\rightarrow$ B $\rightarrow$ C). No hub-and-spoke designs in which every step reads the original input, and no orphan artifacts that no later step reads. The workspace ships the inputs but none of the step outputs, and step~1 must be a real transformation of them. The verifier checks the result of every step and prints one PASS/FAIL line per step.\\[2pt]
\textbf{Difficulty layer.} Common core, plus: at least two edge cases drawn from the skill's own caveats; \emph{inclusion and exclusion}---at least one step must drop items that superficially look processable, and the verifier checks both that the right items are present and that the wrong ones are absent; the difficulty items spread over different steps, with at least one in the second half of the chain.\\[2pt]
\textbf{Self-check.} \emph{``Could a strong agent solve this without the skill's specific knowledge? If yes $\rightarrow$ not hard enough; add a skill-critical step.''} Could the answer be eyeballed or hardcoded from the inputs? Does the verifier recompute expected values?}

\promptbox{Abductive}{box:abductive}{%
\textbf{Fit check.} \emph{``Can this skill host a task with a hidden explanation the solver must infer, and a verifiable check on whether they got it right?''} Otherwise skip (\emph{``not abductive-amenable''}), or skip if the skill offers no uplift anchor.\\[2pt]
\textbf{Task-type contract.} The solver observes evidence and must infer a hidden cause, rule, or explanation, then act on it. Forms include diagnosing a running but wrong system, inducing a rule from input--output examples, reconstructing a cause from logs or corrupted artifacts, and selecting among competing hypotheses. The instruction states the observations and the goal, never the cause or a path to it. The authoritative check stays hidden from the solver and must fail symptom-hiding fixes (swallowing the error, hardcoding the expected output, deleting a failing assertion).\\[2pt]
\textbf{Difficulty layer.} Common core, applied to diagnosis: a realistic multi-module system whose wrong behavior cannot be spotted by reading one file, so the solver must run it and trace the symptom back; verification by recomputation or by invariants on inputs the instruction never enumerates; the planted cause ideally \emph{is} naive code getting a skill-documented caveat wrong. Calibrate the anchoring fact between two failure modes: not general programming knowledge or already present in the workspace (too easy), and not an arbitrary token that exists only in the skill (too hard).\\[2pt]
\textbf{Self-check.} \emph{``Could a generalist spot and fix the defect by reading a few files without the skill? If yes $\rightarrow$ the cause is too shallow.''}}

\promptbox{Constraint satisfaction}{box:csp}{%
\textbf{Fit check.} The skill must offer several competing, deterministically checkable constraints; otherwise skip (\emph{``not CSP-amenable''}) rather than force a checklist into this shape.\\[2pt]
\textbf{Task-type contract.} \emph{``Several simultaneous, individually measurable constraints where the starting state violates at least one and naively fixing one tends to break another.''} The instruction lists the constraints as required outcomes, never how to reconcile them. The verifier checks each constraint as a separate, deterministic, outcome-based assertion and requires all of them.\\[2pt]
\textbf{Difficulty layer.} Common core, plus: \emph{real tension}---at least three constraints with at least one pair where \emph{``the naive fix for constraint A measurably breaks constraint B''}; a setup in which every constraint is independently satisfiable in any order is a checklist and is rejected. No golden configuration: each constraint accepts any configuration that satisfies it. At least one constraint or tension comes from a skill caveat, and at least one satisfying value hinges on a skill-specific threshold, convention, or compatibility rule that the instruction does not restate.\\[2pt]
\textbf{Self-check.} \emph{``Can every constraint be satisfied independently in any order? If yes $\rightarrow$ it is a checklist, not CSP.''} Could a generalist reconcile the constraints without the skill?}

\promptbox{Partial order}{box:partial}{%
\textbf{Fit check.} If the skill's workflow cannot support an edge of the sampled graph, or provides no combining knowledge, skip rather than fake it.\\[2pt]
\textbf{Task-type contract.} The solution is a set of artifact-producing steps \emph{``ordered only where a real dependency demands it---a DAG, not a line.''} For each task a random DAG of 4--6 nodes is sampled and given to the builder as an edge list; every node becomes a concrete artifact and every edge must be real (the consuming step opens the consumed artifact). The graph's motifs force distinct behavior: a \emph{join} combines independently produced artifacts that must agree on a convention (field names, units, ordering, precision); a \emph{skip edge} forces an early artifact to be kept intact until a late consumer; a \emph{fan-out} source must survive all its consumers. At least two intermediate artifacts are \emph{checked}: the verifier confirms that each exists, is correct, and survived (immutable, append-only, or valid), and prints a PASS/FAIL line per node. The instruction may name deliverables but never the combining knowledge or structural vocabulary (``DAG'', ``branch'', ``parallel'').\\[2pt]
\textbf{Difficulty layer.} Common core, plus: \emph{the natural strategy must fail}---a solver that works serially and transforms artifacts in place must fail at least one checked node; a \emph{wrong-by-default primary join}, where the obvious combination yields plausible-looking output that fails on values and the correct combination hinges on skill-provided knowledge that is also recoverable from workspace evidence; and \emph{structure is not the only wall}---\emph{``assume the solver guesses the complete graph''}; the task must still be harder without the skill.\\[2pt]
\textbf{Self-check.} Would an in-place, keep-only-the-latest solver fail a checked node? \emph{``If the solver were handed the full graph, would the task still be much harder without the skill?''}}

\section{Builder and Reviewer Prompts}\label{app:builder_prompts}

Both construction stages (Section~\ref{sec:construction}) pair a builder agent with a reviewer agent. The boxes below condense their system prompts; quoted phrases follow the prompt wording. The task builder's prompt is completed by the task profile of Appendix~\ref{app:profiles}, and the task reviewer's prompt by the profile's reviewer addendum.

\promptbox{Environment builder}{box:env_builder}{%
\textbf{Role.} \emph{``Produce a working Docker image for a single skill so it can later be exercised by a downstream evaluator.''} Tools: read the skill (\texttt{SKILL.md} and its file manifest), read a skill file, build an image from a Dockerfile plus dependency files, and run a script in the built image with the skill mounted read-only.\\[2pt]
\textbf{Procedure.} Read \texttt{SKILL.md} for the language, runtime, package manager, declared dependencies, and system packages. Trust dependency files shipped with the skill (\texttt{requirements.txt}, \texttt{package.json}, \texttt{pyproject.toml}, \ldots); only if a language the skill ships scripts in has none, read a few entry-point scripts and collect their third-party imports. After at most a few reads, build: \emph{``a failed build teaches you more than reading another script.''} On failure, read the build log, find the root cause, and make a targeted edit.\\[2pt]
\textbf{Rules.} Default to \texttt{ubuntu:24.04}, with \texttt{bash} and \texttt{coreutils} available. Declare dependencies in files the builder writes itself and install from those, never from the skill's own copy. Never copy the skill into the image: the skill is mounted at run time, so the image provides dependencies only.}

\promptbox{Environment reviewer}{box:env_reviewer}{%
\textbf{Role.} \emph{``Verify that a previously built Docker image actually has the dependencies the skill needs to run.''} The reviewer reads \texttt{SKILL.md}, then runs one script in the image that checks each declared dependency, and reports success or a precise diagnosis for the rebuild. It verifies but never installs or patches.\\[2pt]
\textbf{Rules.} \emph{``Tests must exercise the dependency, not just inspect it''}: import a package and call something on it, or run a real subcommand of a tool; metadata checks (\texttt{pip show}, \texttt{npm list}, \texttt{which}) are insufficient. If the skill ships scripts, their imports must resolve, which catches dependencies declared nowhere but in the code. Skill files baked into the image are a defect. At most two or three test runs.}

\promptbox{Task builder}{box:task_builder}{%
\textbf{Role.} \emph{``From a single skill and its working environment, author one concrete, automatically-verifiable task that a downstream agent will solve.''} The builder works inside a live container of the skill's environment and can read and run the skill, run shell commands, and create files.\\[2pt]
\textbf{Package.} \texttt{instruction.md} (what the solver reads); \texttt{workspace\_state/} (the solver's starting files); \texttt{solution/solve.sh} (the reference solution); \texttt{tests/test.sh} and helpers (the verifier); and \texttt{grading.json}, which declares the files that carry facts and a checklist of the outputs the verifier grades.\\[2pt]
\textbf{Design principles.} \emph{``Frame the task around a checkable output''}: the deliverable is data the solver produces, checked by value or behavior; a skill whose only possible check is grepping the solver's source is skipped. The workspace is a realistic starting point whose gap to the solution is the work. \emph{``No answers in the workspace''}: no author vocabulary, no comments that explain the hidden cause or rule, no expected values in examples or configs, and no leftovers from the builder's own runs.\\[2pt]
\textbf{Loop.} (1) Read the skill and try its scripts, within a small exploration budget. (2) \emph{Verifiability gate}: \emph{``Can the solver's deliverable be checked by value?''} If not, call \texttt{skip\_task}. (3) Author the package, then call \texttt{validate\_task}, which runs the validity gate and grading checks: the verifier must still pass when fact-carrying files are altered and the reference outputs kept, and must fail when each checklist output is corrupted. Iterate until valid. (4) Call \texttt{review\_task}; on \emph{revise}, fix the verifier (or remove a leak at its source), re-validate, and review again.\\[2pt]
\textbf{Rules.} \emph{``Verify the outcome, not the implementation''}: run the produced artifact or parse its output, and never grep the solver's source for identifiers. \emph{``If a check enforces something the instruction does not already require, drop the check''}; never edit the instruction to justify a check. The instruction uses relative paths only.}

\promptbox{Task reviewer}{box:task_reviewer}{%
\textbf{Role.} \emph{``Judge whether test.sh is a sound reward signal.''} The task has already passed the validity gate, so solvability is not re-examined. The reviewer reads the instruction, the reference solution, and the verifier, and can open any file in the package with read-only tools. \emph{``A sound test is the tightest test that still accepts every correct solution.''}\\[2pt]
\textbf{Defect A: too strict.} \emph{``If the task had been solved a different but valid way, would this check still pass?''} Flags checks on identifiers, imports, formatting, key names, or wording the instruction does not dictate, on one of several allowed options, or on artifacts the instruction never asked for.\\[2pt]
\textbf{Defect B: too weak.} \emph{``Would a lazy or partially-wrong solution still pass?''} Flags existence-only checks, keywords a stub or the starting state already satisfies, structure without values, count thresholds, and multi-step tasks whose intermediate steps are never checked.\\[2pt]
\textbf{Further checks.} Answer leaks in files the solver can see (expected values, thresholds, or the fix in examples, docstrings, comments, or leftover outputs); a declared structure padded to reach its count; an instruction that names the technique or the location of a defect; toy material or a skill used only as a theme.\\[2pt]
\textbf{Verdict.} Identify the task's core outcome and whether the verifier checks it soundly; if so, remove over-strict checks and pass. Otherwise, or on any concrete defect, return \emph{revise} with the file, the line, and the exact fix. The builder has at most five review rounds.}

\section{Example Tasks}\label{app:examples}

We show one task per profile from the \name\ test set, excerpted and lightly formatted. All four are tasks that Kimi-K3 solves with the skill but not without it (Section~\ref{sec:skill_critical}), and that Qwen3.5-9B fails before \name\ SFT and solves after it. Box~\ref{box:ex_procedural} shows a procedural task in more detail; Boxes~\ref{box:ex_abductive}--\ref{box:ex_partial} summarize one task for each of the other profiles.

\promptbox{Procedural task, \texttt{co2lpm} (natural science, held-in skill)}{box:ex_procedural}{%
\textbf{Skill.} Physics knowledge for a 0-D lumped-parameter model of CO$_2$ in geothermal reservoirs: symbols, governing equations, derivations, and sanity checks in four reference files (\texttt{SYMBOLS.md}, \texttt{EQUATIONS.md}, \texttt{DERIVATIONS.md}, \texttt{SANITY\_CHECKS.md}).\\[2pt]
\textbf{Workspace.} \texttt{params.json} (five reservoir scenarios, e.g., \texttt{baseline}, \texttt{reversal}, \texttt{low\_gas\_nodagas}) and \texttt{times.csv} (60 evaluation times from 0 to 100 years).\\[2pt]
\textbf{Instruction} (excerpt). \emph{``The model equations, symbols, derivations, and sanity checks are defined by the skill's reference material \ldots\ Apply those equations exactly as documented. Produce the three deliverables below, in order. Each one feeds the next.''}
(1)~\texttt{derived.json}: per scenario, the time-independent quantities (conductivity $K$, initial and long-term pressure $P_0$, $P_\infty$, critical and effective extraction rates, solubility slope $dC_s/dP$, solubility $C_{s,\infty}$, baseline emissions), \emph{``follow[ing] the solubility convention exactly''}.
(2)~\texttt{trajectory.csv}: read \texttt{derived.json} and compute pressure, upflow, outflow, and solubility at every time, \emph{``using the sign convention stated in the skill.''}
(3)~\texttt{report.json}: read both files, decide whether pressure reversal occurs and when, and integrate the outflow over the 100-year window.\\[2pt]
\textbf{Verifier} (excerpt). Recomputes every expected value from the inputs with an independent implementation and prints one PASS/FAIL line per stage:\\[1pt]
{\ttfamily\tiny
\hspace*{1em}def slope(T): return (A0 + A1*T + A2*T*T) / 1e6 \hfill\# A0, A1, A2 from the skill's solubility law\\
\hspace*{1em}def Cs\_at(dCsdP, Pref, P, degas, C0): return dCsdP*((Pref+P) - P\_SOL\_REF) + C\_00 if degas else C0\\
\hspace*{1em}P = P0 - (q\_eff/K)*(1 - exp(-t/tp)); \ q\_up = -Kup*(Pup - P); \ q\_out = Kout*P\\
\hspace*{1em}rev = q\_eff > q0c; \ t\_r = log(q\_eff/(q\_eff - q0c))*tp if rev else None\\
\hspace*{1em}check("stage1-derived", approx(got[k], exp[k]), ...) \hfill\# relative tolerance 1e-6\par}
\textbf{Why the skill matters.} The instruction names each quantity but never its formula. The solubility-slope coefficients, the degassing rule for $C_s$, the sign convention for upflow, and the reversal-time formula appear only in the skill's reference files, so a solver without the skill must guess them, and any guess fails the recomputed values.}

\promptbox{Abductive task, \texttt{oasis-score} (healthcare, held-in skill)}{box:ex_abductive}{%
\textbf{Instruction} (excerpt). \emph{``The scores in \texttt{patient\_scores.csv} are wrong for some patients. The scoring pipeline completes without errors \ldots\ diagnose why the scores are wrong and fix the root cause so the output is correct for all patients.''} The workspace holds a multi-module pipeline for the OASIS ICU severity score and 80 patient records.\\[2pt]
\textbf{Hidden cause.} Three rule modules (heart rate, mean arterial pressure, temperature) evaluate their bins in the wrong order; because the first matching bin wins, patients with both low and high extremes receive the wrong component score. The instruction only says that bins have a \emph{``defined evaluation order''}; the order itself is given in the skill's MIMIC OASIS tables.\\[2pt]
\textbf{Verifier.} Recomputes the total and component scores of all 80 patients from the specification and compares them with the output of the repaired pipeline.}

\promptbox{Constraint-satisfaction task, \texttt{applying-brand-guidelines} (design, held-out skill)}{box:ex_csp}{%
\textbf{Instruction} (excerpt). \emph{``Edit \texttt{report\_config.json} so that the report configuration satisfies all of the following constraints simultaneously''}: approved color palette, approved font stack, WCAG AA contrast for every text/background pair, standard number and date formats, no prohibited terms, logo size and placement, and table styling. The instruction names the constraints but not the palette, fonts, formats, or logo values.\\[2pt]
\textbf{Tension.} Palette and contrast conflict: several text/background pairs fall below the 4.5:1 ratio (white text on amber reaches only 1.6:1), and each must be repaired using palette colors only, while replacing an off-palette color can in turn break a contrast pair. Four such pairs must be resolved together.\\[2pt]
\textbf{Verifier.} Checks each of the eight constraints separately, recomputing contrast ratios from the final colors, and requires all of them.}

\promptbox{Partial-order task, \texttt{edge-strategy-designer} (finance, held-out skill)}{box:ex_partial}{%
\textbf{Instruction} (excerpt). Turn a batch of trading \emph{edge concepts} into four deliverables: an exit-calibration report, an entry-settings report, one strategy draft per concept variant, and export tickets for the downstream exporter, with \emph{``every calibrated value \ldots\ the one the strategy-design stage's standard rules produce.''}\\[2pt]
\textbf{Graph.} Five nodes: the drafts join the concepts, the exit calibration, and the entry settings, and the tickets reuse the entry settings directly (a skip edge), so the entry report must survive until the last step.\\[2pt]
\textbf{Hidden knowledge.} At the primary join, the obvious choice copies the risk profile's base values (stop loss 0.07, reward-to-risk 3.0) into every draft. This looks valid but is wrong: the skill's script adjusts these values by hypothesis type (e.g., a breakout stop of $0.07\times0.85=0.0595$). A solver without the skill can recover the adjustments only by reverse-engineering example drafts from a previous run in the workspace.\\[2pt]
\textbf{Verifier.} Recomputes the calibrated values with the skill's rules and checks the two reports, every draft, and every ticket, printing one PASS/FAIL line per node.}

\section{Additional Reference Models}\label{app:reference_models}

Table~\ref{tab:reference_models} reports all reference models evaluated under the same protocol as Table~\ref{tab:main_experiments}.

\begin{table}[h]
\centering
\caption{All reference models, with skills available, evaluated under the same protocol as Table~\ref{tab:main_experiments}. $^{*}$\name\ score over the tasks graded so far. $^{\diamond}$Teacher models.}
\label{tab:reference_models}
\small
\setlength{\tabcolsep}{3pt}
\renewcommand{\arraystretch}{1.15}
\begin{tabular*}{\textwidth}{@{\extracolsep{\fill}}lrcccc@{}}
\toprule
\textbf{Model} & \textbf{Size} & \textbf{\name} & \textbf{SkillEval} & \textbf{SkillsBench} & \textbf{Skill-Use-Bench} \\
\midrule
Qwen3.5 & 397B/17B & $55.8$ & $73.8 \pm 0.8$ & $33.2 \pm 3.0$ & $23.0$ \\
\addlinespace[2pt]
Qwen3.8 & 27B & $69.0$ & $81.6 \pm 0.8$ & $65.0 \pm 1.5$ & $68.0$ \\
\addlinespace[2pt]
GLM-5.2$^{\diamond}$ & 753B & $66.5$ & $79.6 \pm 0.8$ & $57.8 \pm 4.8$ & $76.3$ \\
GLM-5.3 & 753B & $68.8$ & $80.8 \pm 0.7$ & $56.9 \pm 2.0$ & $78.2$ \\
\addlinespace[2pt]
DeepSeek-V4-Pro & 1.6T/49B & $61.5$ & $78.8 \pm 0.1$ & $57.0 \pm 2.6$ & $57.7$ \\
DeepSeek-V4-Flash$^{\diamond}$ & 284B/13B & $61.5$ & $78.6 \pm 0.2$ & $58.7 \pm 1.0$ & $56.1$ \\
\addlinespace[2pt]
MiniMax-M3 & 428B/23B & $64.5$ & $79.1 \pm 1.4$ & $53.4 \pm 2.0$ & $72.3$ \\
\bottomrule
\end{tabular*}
\end{table}

\section{Dataset Statistics}\label{app:statistics}

Table~\ref{tab:stats_funnel} traces skills from crawling to selection, Table~\ref{tab:stats_domains} gives the domain distribution of selected skills and constructed tasks, Table~\ref{tab:stats_tasks} summarizes the tasks, and Table~\ref{tab:stats_traj} summarizes the trajectories used for training.

\begin{table}[h]
\centering
\caption{Skill curation funnel by source. Registry skills are fetched from their GitHub repositories; skills that could not be fetched and exact duplicates are removed before screening. The last row removes skills that appear in both sources.}
\label{tab:stats_funnel}
\small
\begin{tabular}{@{}lrrr@{}}
\toprule
\textbf{Stage} & \textbf{skills.sh} & \textbf{Registry} & \textbf{Total} \\
\midrule
Crawled (manifest) & 9,724 & 174,706 & 184,430 \\
Fetched and deduplicated & 8,593 & 42,538 & 51,131 \\
Screened as a real skill & 8,104 & 37,333 & 45,437 \\
Annotated (not archived or bulk-published) & 6,776 & 20,793 & 27,569 \\
Selected (rules in Table~\ref{tab:skill_annotation}) & 2,838 & 9,268 & 12,106 \\
After cross-source deduplication & 2,838 & 9,059 & 11,897 \\
\bottomrule
\end{tabular}
\end{table}

\begin{table}[h]
\centering
\caption{Domain distribution of the selected skill pool and of the constructed tasks. Task construction samples skills with per-domain quotas, which cap software engineering and exclude agents/meta and writing skills, so the task distribution is more balanced than the pool.}
\label{tab:stats_domains}
\small
\begin{tabular}{@{}lrrr@{}}
\toprule
\textbf{Domain} & \textbf{Selected skills} & \textbf{Skills with tasks} & \textbf{Tasks} \\
\midrule
Marketing & 406 & 374 & 798 \\
Software engineering & 7,726 & 707 & 791 \\
Productivity & 321 & 285 & 618 \\
Cybersecurity & 341 & 289 & 602 \\
ML / AI tooling & 286 & 263 & 598 \\
Infrastructure & 317 & 283 & 594 \\
Data analytics & 286 & 249 & 531 \\
Media & 242 & 221 & 447 \\
Office & 234 & 201 & 405 \\
Crypto / Web3 & 121 & 116 & 257 \\
Finance & 105 & 97 & 225 \\
Healthcare & 75 & 71 & 167 \\
Communication & 84 & 79 & 151 \\
Natural science & 68 & 59 & 137 \\
Mathematics & 32 & 29 & 74 \\
Robotics & 38 & 34 & 66 \\
Manufacturing & 24 & 21 & 46 \\
Energy & 6 & 5 & 10 \\
Agents / meta & 651 & 0 & 0 \\
Writing & 414 & 0 & 0 \\
Other & 120 & 111 & 255 \\
\midrule
Total & 11,897 & 3,494 & 6,772 \\
\bottomrule
\end{tabular}
\end{table}

\begin{table}[h]
\centering
\caption{Constructed tasks. Top: counts by reasoning profile and review outcome (\emph{passed}: validity gate and reviewer approval; \emph{unresolved}: valid, but the reviewer's strictness concerns were not fully resolved within the round limit). Partial-order tasks were built in separate construction runs and are not included in these counts. Bottom: task size over the 6,692 locally available packages (median, mean, and 10th--90th percentile).}
\label{tab:stats_tasks}
\small
\begin{tabular}{@{}lrrr@{}}
\toprule
\textbf{Profile} & \textbf{Passed} & \textbf{Unresolved} & \textbf{Total} \\
\midrule
Procedural execution & 1,293 & 869 & 2,162 \\
Abductive diagnosis & 2,400 & 430 & 2,830 \\
Constraint satisfaction & 1,498 & 282 & 1,780 \\
\midrule
Total & 5,191 & 1,581 & 6,772 \\
\midrule
\textbf{Size} & \textbf{Median} & \textbf{Mean} & \textbf{P10--P90} \\
\midrule
Instruction length (words) & 336 & 365.4 & 215--556 \\
Workspace files & 9 & 25.1 & 1--30 \\
Workspace size (KiB) & 19 & 1,261.2 & 2--99 \\
Verifier code (lines) & 281 & 305.3 & 148--494 \\
\bottomrule
\end{tabular}
\end{table}

\begin{table}[h]
\centering
\caption{Collected trajectories. Successful trajectories pass the task verifier; unsuccessful ones are graded and fail. Length statistics give the median (mean) per trajectory; Terminus-2 issues commands inside its JSON reply rather than as tool calls.}
\label{tab:stats_traj}
\small
\begin{tabular}{@{}llrr@{}}
\toprule
& & \textbf{Successful} & \textbf{Unsuccessful} \\
\midrule
\multicolumn{2}{@{}l}{Trajectories} & 19,070 & 15,649 \\
\midrule
\multirow{3}{*}{Harness}
& MiniSwe-Agent & 6,433 & 4,360 \\
& AgentFly & 5,936 & 6,909 \\
& Terminus-2 & 6,701 & 4,380 \\
\midrule
\multirow{3}{*}{Teacher}
& Kimi-K3 & 10,603 & 7,649 \\
& DeepSeek-V4-Flash & 6,521 & 5,507 \\
& GLM-5.2 & 1,946 & 2,493 \\
\midrule
\multirow{4}{*}{Length}
& Tokens & 25,146 (29,254.2) & 35,977 (43,157.5) \\
& Reasoning tokens & 7,008 (9,952.2) & 12,460 (17,854.6) \\
& Assistant turns & 12 (13.9) & 15 (17.9) \\
& Tool calls & 14 (14.6) & 18 (18.9) \\
\bottomrule
\end{tabular}
\end{table}

\section{Task Structure Annotation}\label{app:task_structure}

\begin{figure}
\centering
\includegraphics[width=\textwidth]{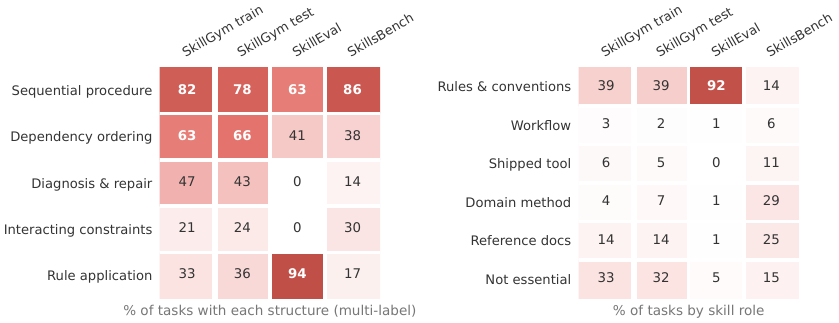}
\caption{Share of tasks annotated with each reasoning structure (left; a task can have several) and each skill role (right). Columns: a stratified \name\ training sample ($n{=}300$), the \name\ test set ($n{=}400$), SkillEval ($n{=}100$), and SkillsBench ($n{=}87$).}
\label{fig:task_composition}
\end{figure}

\begin{figure}
\centering
\includegraphics[width=\textwidth]{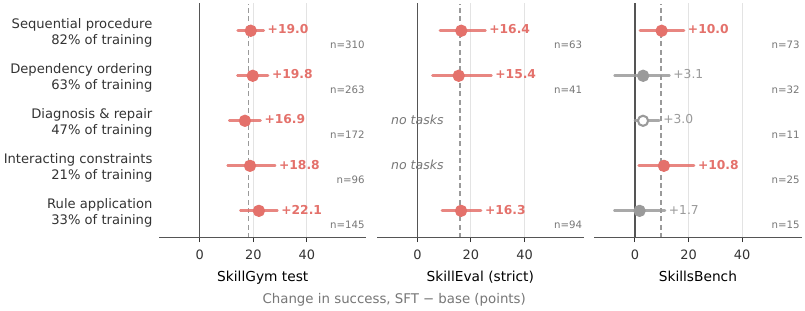}
\caption{Change in success from Qwen3.5-9B base to \name\ SFT for tasks with each structure, with skills available. Dots show mean paired differences and lines 95\% task-bootstrap intervals; coral marks intervals that exclude zero, and hollow markers denote fewer than 15 tasks. Dashed lines show each set's overall gain. Row labels give each structure's share of the training sample. SkillEval uses strict success averaged over three runs; SkillsBench uses reward averaged over three runs, excluding trials lost to an environment permission failure.}
\label{fig:gains_by_structure}
\end{figure}

\paragraph{Rubric.} Each task is labeled on three facets. \emph{Reasoning structure} (multi-label): sequential procedure ($\geq$3 dependent steps in which a checked output depends on an intermediate result); dependency ordering ($\geq$2 checked deliverables whose dependencies form a graph rather than a chain); diagnosis and repair (an unstated defect must be located and fixed, and the verifier checks corrected behavior); interacting constraints ($\geq$2 checked constraints where satisfying one can violate another); and rule application (rules, thresholds, or conventions given in the skill applied to each item of a collection). An additional \emph{other} label requires a free-text description; it was used once among 887 tasks. \emph{Skill role} (single label): rules and conventions, workflow, shipped tool, domain method, reference documentation, not essential, or other. \emph{Action} (single label): read and answer, compute over inputs, modify an existing system, or generate many artifacts. Two flags record leakage of the intended reasoning in solver-visible material and verifier checks of conventions that neither the instruction nor the skill determines. For every positive label, the annotator must quote supporting evidence from the task, and it reports its confidence for each facet.

\paragraph{Inputs and model.} The annotator sees the instruction, a listing and previews of the initial workspace, the skill documents, and the verifier code. Reference solutions, design notes, and task metadata are withheld, and construction-profile identifiers are redacted from every set. We use GPT-5.4, which did not construct or review \name\ tasks. Replies that violate the schema are returned to the model with the validation error. As a check of the verifier reading, the annotated number of checks matches the exact count in SkillEval's evaluator specification for 99 of 100 tasks.

\paragraph{Gain analysis.} Outcomes are paired per task: \name\ success from one run; SkillEval strict success averaged over three runs; SkillsBench reward averaged over the runs in which the task was graded, excluding trials lost to an environment permission failure. These per-task outcomes reproduce the aggregates in Table~\ref{tab:main_experiments}. Intervals are 95\% bootstrap intervals over tasks (2{,}000 resamples). Table~\ref{tab:structure_regression} regresses the paired difference on the structure labels, the evaluation set, and difficulty, defined as the mean outcome of nine reference models on the task (excluding the base and SFT models).

\begin{table}[h]
\centering
\caption{Regression of the paired SFT$-$base difference (points) on task properties ($n{=}585$ tasks). Intervals are 95\% bootstrap intervals over tasks (1{,}000 resamples). Set effects are relative to \name; difficulty ranges from 0 to 1.}
\label{tab:structure_regression}
\small
\begin{tabular}{lrc}
\toprule
Term & Coefficient & 95\% CI \\
\midrule
Sequential procedure & $+4.6$ & $[-4.2, +12.5]$ \\
Dependency ordering & $+0.4$ & $[-6.5, +7.2]$ \\
Diagnosis and repair & $-2.3$ & $[-9.8, +5.6]$ \\
Interacting constraints & $+1.0$ & $[-8.7, +11.2]$ \\
Rule application & $+6.3$ & $[-2.0, +15.0]$ \\
SkillEval & $-1.2$ & $[-11.7, +9.2]$ \\
SkillsBench & $-6.2$ & $[-14.7, +3.2]$ \\
Difficulty (reference success) & $+18.0$ & $[+11.0, +25.6]$ \\
Intercept & $+1.7$ & $[-8.3, +11.8]$ \\
\bottomrule
\end{tabular}
\end{table}

\section{Cost Estimation}\label{app:cost}

We estimate the API cost of building \name\ at DeepSeek's off-peak prices, pricing task construction at DeepSeek-V4-Pro and trajectory collection at DeepSeek-V4-Flash. Since agent loops re-send the conversation history, we split input tokens into cached tokens $T_{\mathrm{hit}}$ and new tokens $T_{\mathrm{miss}}$, and the cost of a stage with output tokens $T_{\mathrm{out}}$ and prices $p$ per million tokens is
$$
  C = \left(T_{\mathrm{hit}}\,p_{\mathrm{hit}} + T_{\mathrm{miss}}\,p_{\mathrm{miss}} + T_{\mathrm{out}}\,p_{\mathrm{out}}\right)/10^{6}.
$$
Token counts come from the saved builder transcripts and a sample of 3{,}000 trajectories. Table~\ref{tab:cost} shows that task construction dominates the cost, including failed and skipped attempts. Reviewer transcripts were not stored, so the review cost assumes 12k input and 15k output tokens per call. In total, \name\ costs \$5.7k--8.4k, or about \$1 per released task, excluding environment building, annotation, and evaluation.

\begin{table}[h]
\centering
\caption{Estimated API cost of building \name\ at DeepSeek off-peak prices per million tokens, with \$0.022, \$0.66, and \$1.98 for cached input, new input, and output on DeepSeek-V4-Pro, and \$0.003, \$0.15, and \$0.60 on DeepSeek-V4-Flash. Token counts are in billions.}
\label{tab:cost}
\small
\begin{tabular}{llrrrr}
\toprule
Stage & Priced as & Cached in & New in & Output & Cost (USD) \\
\midrule
Task construction, measured & V4-Pro & 58.4 & 0.54 & 1.00 & 3.6k \\
Task construction, all attempts & V4-Pro & \multicolumn{3}{c}{extrapolated} & up to 6.3k \\
Quality review, assumed & V4-Pro & 0.28 & 0.39 & 0.84 & 1.9k \\
Trajectory collection & V4-Flash & 6.44 & 0.48 & 0.08 & 0.14k \\
\midrule
Total & & & & & 5.7k--8.4k \\
\bottomrule
\end{tabular}
\end{table}

\end{document}